\documentclass[journal]{IEEEtran}
\usepackage[nocompress]{cite}
\usepackage[pdftex]{graphicx}
\newcommand{\etal}{\mbox{\emph{et al.\ }}}
\newcommand{\ie}{\mbox{\emph{i.e.,}}}

\usepackage{amsmath}
\usepackage{multirow}

\usepackage{amssymb}

\usepackage{algorithmic}
\usepackage{array}
\begin{document}

\title{A Discriminatively Learned CNN Embedding for Person Re-identification}

\author{Zhedong Zheng, Liang Zheng and Yi Yang \thanks{Zhedong Zheng, Liang Zheng and Yi Yang are with Faculty of Engineering and IT, University of Technology Sydney, NSW, Australia. E-mail: zdzheng12@gmail.com, liangzheng06@gmail.com, yee.i.yang@gmail.com}}

% The paper headers
\markboth{Journal of \LaTeX\ Class Files,~Vol.~14, No.~8, August~2015}%
{Shell \MakeLowercase{\textit{et al.}}: A Discriminatively Learned CNN Embedding for Person Re-identification}
% The only time the second header will appear is for the odd numbered pages
% after the title page when using the twoside option.
% 
% *** Note that you probably will NOT want to include the author's ***
% *** name in the headers of peer review papers.                   ***
% You can use \ifCLASSOPTIONpeerreview for conditional compilation here if
% you desire.

% If you want to put a publisher's ID mark on the page you can do it like
% this:
%\IEEEpubid{0000--0000/00\$00.00~\copyright~2015 IEEE}
% Remember, if you use this you must call \IEEEpubidadjcol in the second
% column for its text to clear the IEEEpubid mark.

% use for special paper notices
%\IEEEspecialpapernotice{(Invited Paper)}

% make the title area
\maketitle

% As a general rule, do not put math, special symbols or citations
% in the abstract or keywords.
\begin{abstract}
In this paper, we revisit two popular convolutional neural networks (CNN) in person re-identification (re-ID), \ie verification and identification models. The two models have their respective advantages and limitations due to different loss functions. In this paper, we shed light on how to combine the two models to learn more discriminative pedestrian descriptors. Specifically, we propose a siamese network that simultaneously computes the identification loss and verification loss. Given a pair of training images, the network predicts the identities of the two input images and whether they belong to the same identity. Our network learns a discriminative embedding and a similarity measurement at the same time, thus making full usage of the re-ID annotations. 

Our method can be easily applied on different pre-trained networks. Albeit simple, the learned embedding improves the state-of-the-art performance on two public person re-ID benchmarks. Further, we show our architecture can also be applied in image retrieval.
\end{abstract}

% Note that keywords are not normally used for peerreview papers.

% For peer review papers, you can put extra information on the cover
% page as needed:
% \ifCLASSOPTIONpeerreview
% \begin{center} \bfseries EDICS Category: 3-BBND \end{center}
% \fi
%
% For peerreview papers, this IEEEtran command inserts a page break and
% creates the second title. It will be ignored for other modes.
\IEEEpeerreviewmaketitle

\section{Introduction}
% The very first letter is a 2 line initial drop letter followed
% by the rest of the first word in caps.
% 
% form to use if the first word consists of a single letter:
% \IEEEPARstart{A}{demo} file is ....
% 
% form to use if you need the single drop letter followed by
% normal text (unknown if ever used by the IEEE):
% \IEEEPARstart{A}{}demo file is ....
% 
% Some journals put the first two words in caps:
% \IEEEPARstart{T}{his demo} file is ....
% 
% Here we have the typical use of a "T" for an initial drop letter
% and "HIS" in caps to complete the first word.

\IEEEPARstart{P}erson re-identification (re-ID) is usually viewed as an image retrieval problem, which matches pedestrians from different cameras \cite{zheng2016survey}. Given a person-of-interest (query), person re-ID determines whether the person has been observed by another camera. Recent progress in this area has been due to two factors: 1) the availability of the large-scale pedestrian datasets. The datasets contain the general visual variance of pedestrian and provide a comprehensive evaluation \cite{li2014deepreid,zheng2015scalable}. 2) the learned embedding of pedestrian using a convolutional neural network (CNN).
%Some previous researches focus on designing good hand-crafted feature and learning similarity metric to solve this problem\cite{yang2014salient,liao2015person,pedagadi2013local,liu2014semi}. 

\begin{figure}[t]
\begin{center}
%\fbox{\rule{0pt}{2in} \rule{0.9\linewidth}{0pt}}
   \includegraphics[width=0.9\linewidth]{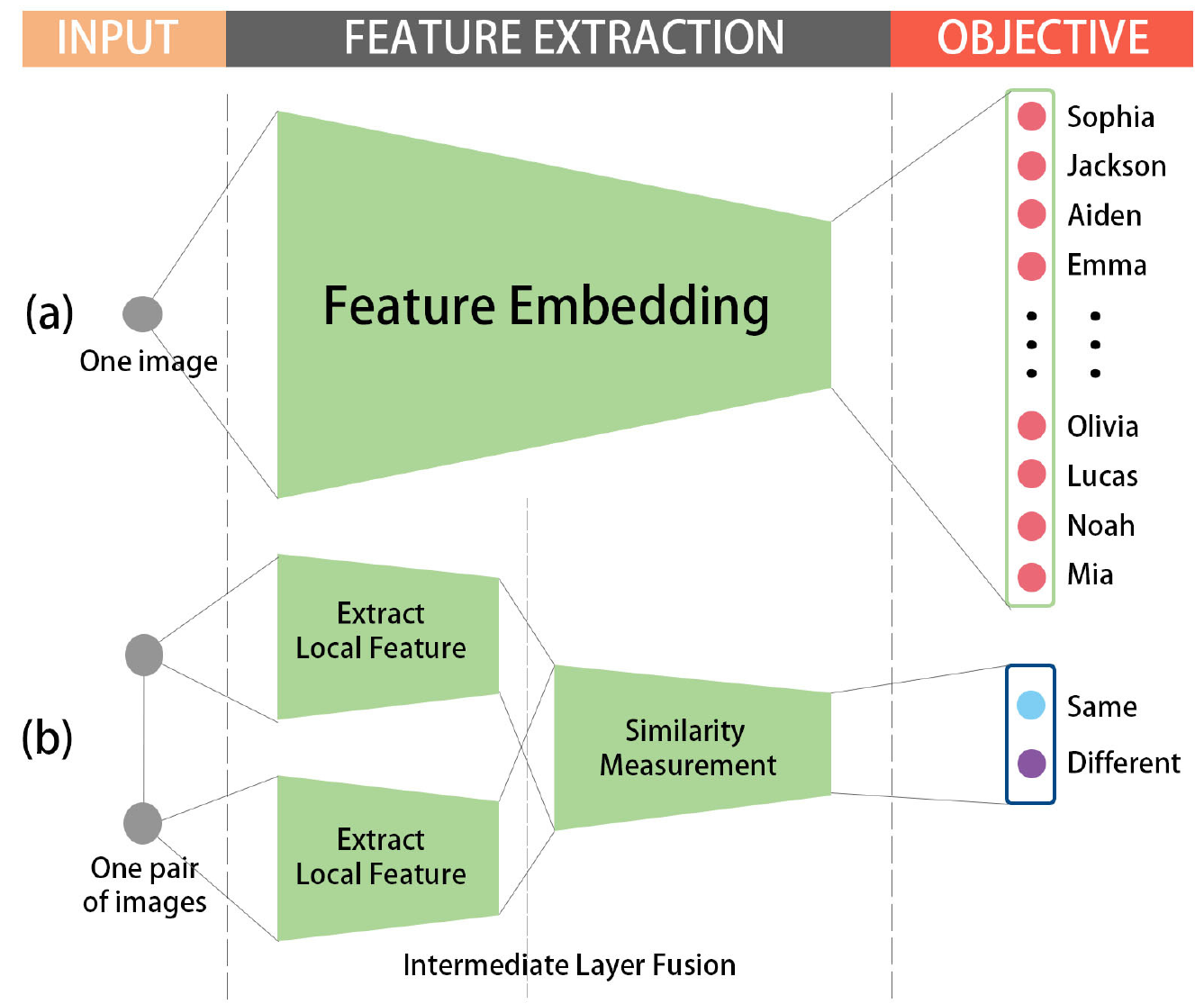}
\end{center}
   \caption{ The difference between the verification and identification models. Green blocks represent non-linear functions by CNN.
   a) Identification models treat person re-ID as a multi-class recognition task, which take one image as input and predict its identity.
   b) Verification models treat person re-ID as a two-class recognition task or a similarity regression task, which take a pair of images as input and determine whether they belong to the same person or not. Here we only show a two-class recognition case.}
\label{fig:1}
\end{figure}

Recently, the convolutional neural network (CNN) has shown potential for learning state-of-the-art feature embeddings or deep metrics \cite{li2014deepreid,yi2014deep,wu2016personnet,varior2016gated,xiao2016learning,zheng2016person,xiao2016end}. As shown in Fig. \ref{fig:1}, there are two major types of CNN structures, \ie verification models and identification models. The two models are different in terms of input, feature extraction and loss function for training. Our motivation is to combine the strengths of the two models and learn a more discriminative pedestrian embedding.
 
Verification models take a pair of images as input and determine whether they belong to the same person or not. A number of previous works treat person re-ID as a binary-class classification task or a similarity regression task \cite{li2014deepreid,yi2014deep,wu2016personnet,varior2016gated}. Given a label $s\in\{0,1\}$, the verification network forces two images of the same person to be mapped to nearby points in the feature space. If the images are of different people, the points are far apart. However, the major problem in the verification models is that they only use weak re-ID labels \cite{zheng2016survey}, and do not take all the annotated information into consideration. Therefore, the verification network lacks the consideration of the relationship between the image pairs and other images in the dataset. 

In the attempt to take full advantages of the re-ID labels, identification models which treat person re-identification as a multi-class recognition task, are employed for feature learning \cite{xiao2016learning,zheng2016person,zheng2016survey,xiao2016end}. They directly learn the non-linear functions from an input image to the person ID and the cross-entropy loss is used following the final layer. During testing, the feature is extracted from a fully connected layer and then normalized. The similarity of two images is thus computed by the Euclidean distance between their normalized CNN embeddings. The major drawback of the identification model is that the training objective is different from the testing procedure, \ie it does not account for the similarity measurement between image pairs, which can be problematic during the pedestrian retrieval process. 

\begin{table}
\begin{center}
\begin{tabular}{c|ccc}
\hline
\multirow{2}{*}{Method} & \multirow{2}{*}{Strong Label} & Similarity& Re-ID\\
 & & Estimation & Performance\\
\hline
Verification Models& $\times$ & \checkmark & fair \\
%\hline
Identification Models& \checkmark & $\times$ & good\\
%\hline
Our Model & \checkmark& \checkmark & good\\
\hline
\end{tabular}
\end{center}
\caption{The advantages and disadvantages of verification and identification models are listed. We assume sufficient training data in all models. Our model takes the advantages of the two models.}
\label{table:comparison}
\end{table}

The above-mentioned observations demonstrate that the two types of models have complementary advantages and limitations as shown in Table \ref{table:comparison}. Motivated by these properties, this work proposes to combine the strengths of the two networks and leverage their complementary nature to improve the discriminative ability of the learned embeddings. The proposed model is a siamese network that predicts person identities and similarity scores at the same time. Compared to previous networks, we take full advantages of the annotated data in terms of pair-wise similarity and image identities. During testing, the final convolutional activations are extracted for Euclidepdfan distance based pedestrian retrieval.
%Our siamese network is designed to share filter's weight. So when extracting feature, our network is degrade to the classification model, which accelerates the testing without performance loss. 
To summarize, our contributions are:
%Fig. 3 illustrates a sample batch of size $m = 5$ for the verification model and our model. The number in the circle is the identity label. Red line means the image pair depicts the different identities. And blue line means the pair depicts the same identity. In contrast, we explicitly take more relationship between data points into consideration.

\begin{itemize}
\item We propose a siamese network that has two losses: identification loss and verification loss. This network simultaneously learns a discriminative CNN embedding and a similarity metric, thus improving pedestrian retrieval accuracy. 
%\item Our network is efficient as classification model when extracting feature.
\item We report competitive accuracy compared to the state-of-art methods on two large-scale person re-ID datasets (Market1501 \cite{zheng2015scalable} and CUHK03 \cite{li2014deepreid}) and one instance retrieval dataset (Oxford5k \cite{philbin2007object}).  
\end{itemize}

The paper is organized as follows. We first review some related works in Section \ref{sec:related_work}.
In Section \ref{method}, we describe how we combine the two losses and define the CNN structure. The implementation details are provided. In Section \ref{exp}, we present the experimental results on two large-scale person re-identification datasets and one instance retrieval dataset. We conclude this paper in Section \ref{sec:conclusion}.

\section{Related Work}\label{sec:related_work}
In this section we describe previous works relevant to the approach discussed in this paper.
They are mainly based on verification models or identification models.

\subsection{Verification Models}
In 1993, Bromley \etal \cite{bromley1993signature} first used verification models to deep metric learning in signature verification. Verification models usually take a pair of images as input and output a similarity score by calculating the cosine distance between low-dimensional features, which can be penalized by the contrastive loss. Recently researchers have begun to apply verification models to person re-identification with a focus on data augmentation and image matching. 
Yi \etal \cite{yi2014deep} split a pedestrian image into three horizontal parts and train three part-CNNs to extract features. The similarity of two images is computed by the cosine distance of their features. Similarly, Cheng \etal split the convolutional map into four parts and fuse the part features with the global features \cite{cheng2016person}.
Li \etal \cite{li2014deepreid} add a patch-matching layer that multiplies the activation of two images in different horizontal stripes. They use it to find similar locations and treat similarity regression as binary-class  penalized by softmax loss. 
Later, Ahmed \etal \cite{ahmed2015improved} improve the verification model by adding a different matching layer that compares the activation of two images in neighboring pixels.
Besides, Wu \etal \cite{wu2016personnet} use smaller filters and a deeper network to extract features. 
Varior \etal \cite{varior2016gated} combine CNN with some gate functions, similar to long-short-term memory (LSTM \cite{hochreiter1997long}) in spirit, which aims to adaptively focus on the similar parts of input image pairs. But it is limited by the computational inefficiency because the query image has to pair with every gallery image to pass through the network. Moreover, Ding \etal \cite{ding2015deep} use triplet samples for training the network which considers the images from the same people and the different people at the same time. 

\subsection{Identification Models}
Recent datasets such as CUHK03 \cite{li2014deepreid} and Market1501 \cite{zheng2015scalable} provide large-scale training sets, which make it possible to train a deeper classification model without over-fitting. Every identity has 9.6 training images on average in CUHK03 \cite{li2014deepreid} and has 17.2 images in Market1501 \cite{zheng2015scalable}. CNN can learn discriminative embeddings by itself without part-matching. Zheng \etal \cite{zheng2016survey,zheng2016person,zheng2016mars} directly use a conventional fine-tuning approach on Market1501 \cite{zheng2015scalable}, PRW \cite{zheng2016person} and MARS \cite{zheng2016mars} and outperform many recent results. 
Wu \etal \cite{wu2016enhanced} combine CNN embeddings with the hand-crafted features in the FC layer.
Besides, Xiao \etal \cite{xiao2016learning} jointly train a classification model using multiple datasets and propose a new dropout function to deal with the hundreds of classes. In \cite{xiao2016end}, Xiao \etal train a classification model similar to the faster-RCNN \cite{girshick2015fast} method and automatically predict the location of the candidate pedestrian from the whole image, which alleviates the pedestrian detection errors. 

\begin{figure}[t]
\begin{center}
%\fbox{\rule{0pt}{2in} \rule{0.9\linewidth}{0pt}}
   \includegraphics[width=1\linewidth]{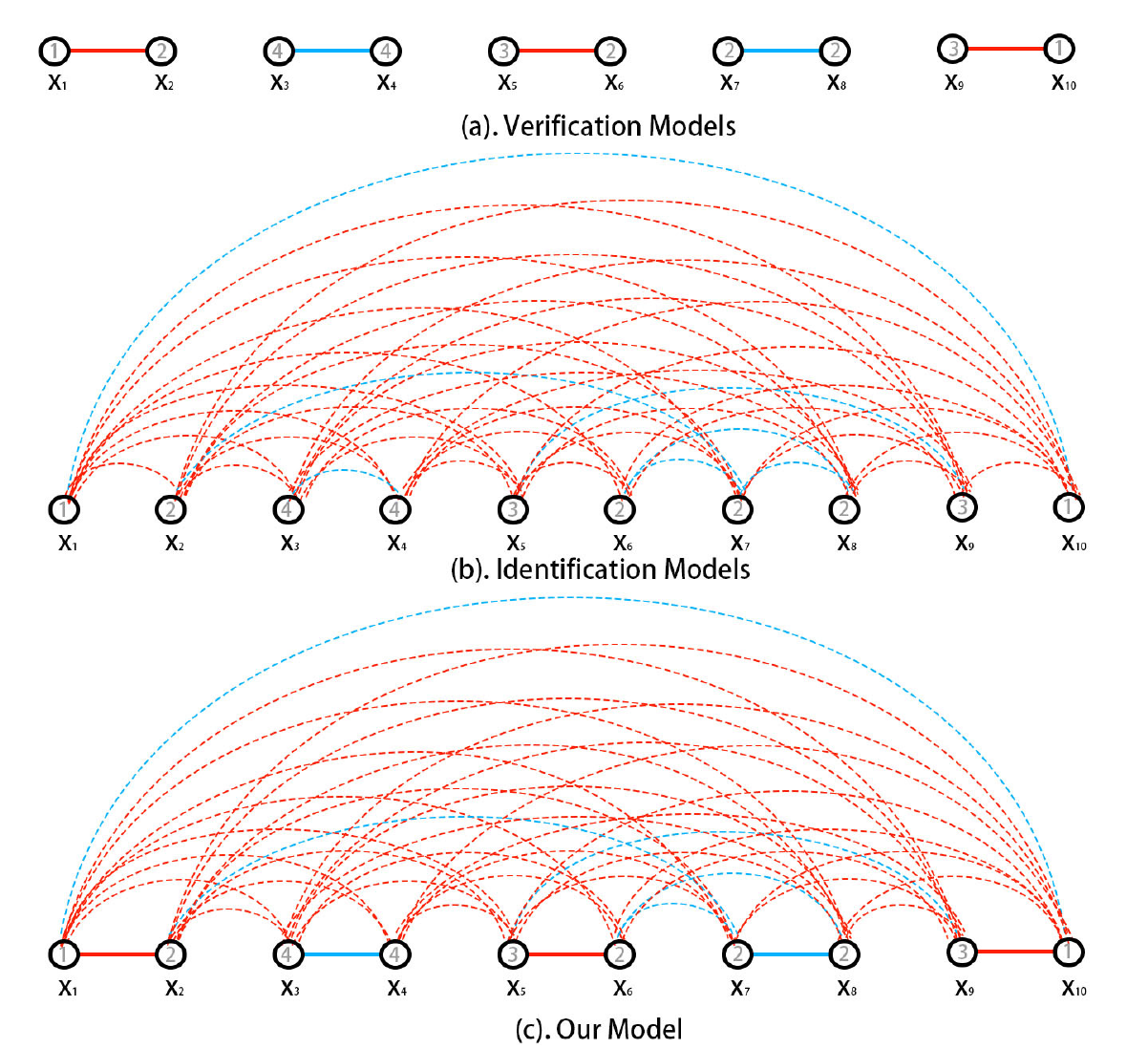}
\end{center}
   \caption{Illustration for a training batch. The number in the circle is the identity label. Blue and red edges represent whether the image pair depicts the same identity or not. Dotted edges represent implicit relationships and solid edges represent explicit relationships. Our model combine the strengths of the two models.}
\label{fig:3}
\label{fig:onecol}
\end{figure}

\begin{figure*}
\begin{center} 
%\fbox{\rule{0pt}{2in} \rule{.9\linewidth}{figure2.pdf}}
\includegraphics[width=1\linewidth]{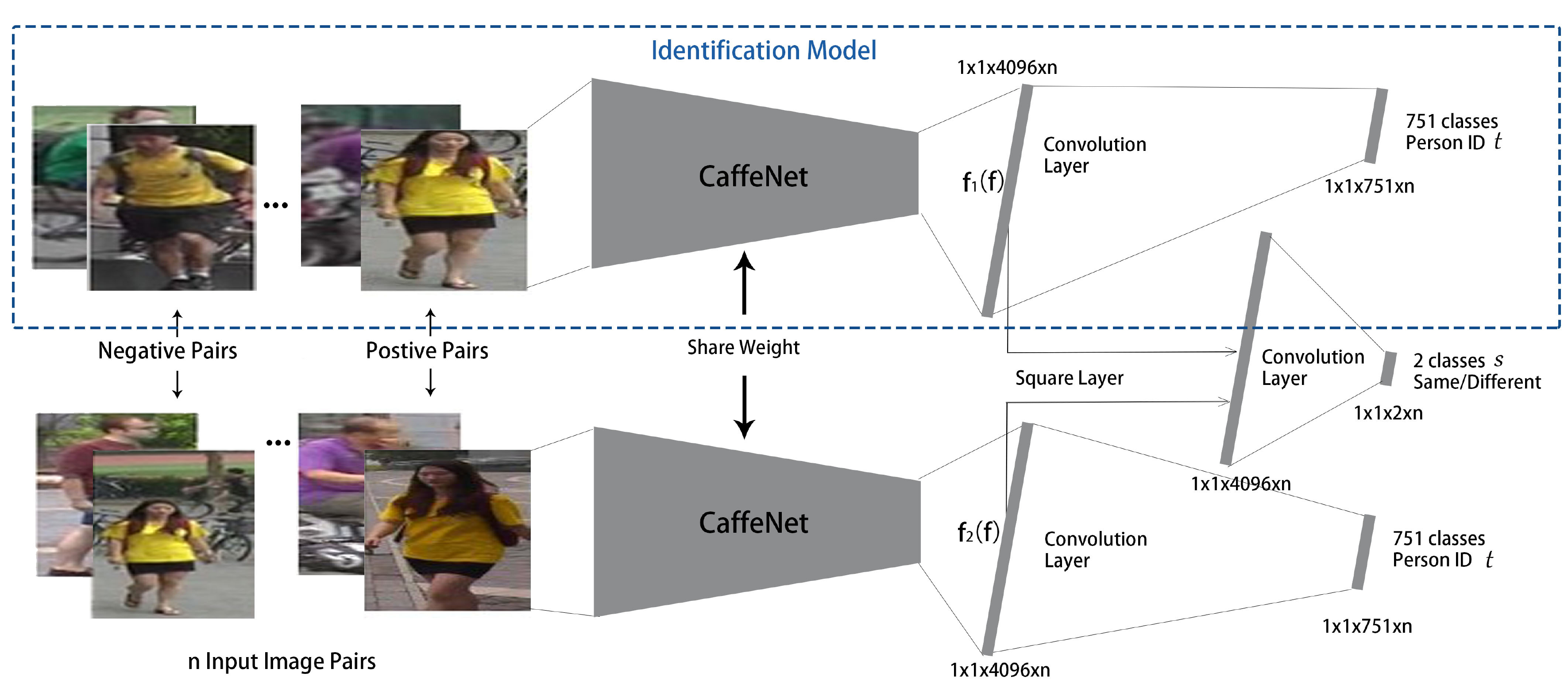}
\end{center}
   \caption{The proposed model structure. Given $n$ pairs of images of size $227\times227$, two identical CaffeNet models are used as the non-linear embedding functions and output 4,096-dim embeddings $f_1,f_2$. Then, $f_1,f_2$ are used to predict the identity $t$ of the two input images, respectively, and also predict the verification label $s$ jointly. We introduce a non-parametric layer called Square Layer to compare high level features $f_1,f_2$. Finally, the softmax loss is applied on the three objectives.}
\label{fig:2}
\end{figure*}
\subsection{Verification-identification Models}
In face recognition, the ``DeepID networks'' train the network with the verification and identification losses \cite{sun2014deep,sun2015deeply,sun2015deepid3}, which is similar to our network. In \cite{sun2014deep}, Sun \etal jointly train face identification and verification. Then more verification supervision is added into the model \cite{sun2015deeply} and a deeper network is used \cite{sun2015deepid3}. 

Our method is different from their models in the following aspects. First, in face recognition, the training dataset contains 202,599 face images of 10,177 identities \cite{sun2014deep} while the current largest person re-id training dataset contains 12,936 images of 751 identities \cite{zheng2015scalable}. DeepID networks apply contrastive loss to the verification problem, wile our model uses the cross-entropy loss. We find that the contrastive loss leads to over-fitting when the number of images is limited. In the experiment, we show the proposed method learns more robust person representative and outperforms using contrastive loss. Second, dropout \cite{srivastava2014dropout} cannot be applied on the embedding before the contrastive loss, which introduces zero values at random locations. On the contrary, we can add dropout regularization on the embedding in the proposed model. Third, the DeepID networks are trained from scratch, while our model benefits from the networks pretrained on ImageNet \cite{russakovsky2015imagenet}. Finally, we evaluate our method on the tasks of person re-ID and instance retrieval, providing more insights in the verification-classification models. 

%Compared with these methods, our model benefits from considering verification and identification losses at the same time. We optimize the algorithm based on the person re-identification problem. 

\section{Proposed Method} \label{method}

\subsection{Preview}
Fig. \ref{fig:3} (a) and Fig. \ref{fig:3} (b) illustrate the relational graph built by verification and identification models. In a sample batch of size $m = 10$, red edges represent the positive pairs (the same person) and blue edges represent the negative pairs (different persons). The dotted edges denote implicit relationships built by the identification loss and the solid edges denote explicit relationships built by the verification loss. 

In verification models, there are several operations between the two inputs. The explicit relationship between data is built by the pair-wise comparison, such as part matching \cite{li2014deepreid,ahmed2015improved} or contrastive loss \cite{hadsell2006dimensionality}. For example, contrastive loss directly calculates the Euclidean distance between two embeddings. In identification models, the input is independent to each other. But there is implicit relationship between the learned embeddings built by the cross-entropy loss. The cross-entropy loss can be formulated as $loss=-log(p_{gt}), \mbox{where } p_{gt} = W_{gt}f_i$. $W$ is the weight of the linear function. $f_m, f_n$ are the embeddings of the two images $x_m, x_n$ from the same class $k$.  To maximize $W_{k}f_{m}$, $W_{k}f_{n}$, the network converges when $f_m$ and $f_n$ have similar vector direction with $W_k$. In \cite{liu2016large}, similar observation and visualization are shown. So the learned embeddings are eventually close for images within the same class and far away for images in  the different classes. The relationship is implicitly built between $x_m, x_n$ and bridged by the weight $W_k$.

Due to the usage of the weak labels, verification models take limited relationships into consideration. On the other hand, classification models do not explicitly consider similarity measurements. Fig. \ref{fig:3} (c) illustrates how our model works in a batch. We benefit from simultaneously considering the verification and identification losses. The proposed model thus combines the strength of the two models (see Table \ref{table:comparison}).

\subsection{Overall Network}
Our network is basically a convolutional siamese network that combines the verification and identification losses. Fig. \ref{fig:2} briefly illustrates the architecture of the proposed network. Given an input pair of images resized to $227\times227$, the proposed network simultaneously predicts the IDs of the two images and the similarity score. The network consists of two ImageNet \cite{russakovsky2015imagenet} pre-trained CNN models, three additional Convolutional Layers, one Square Layer and three losses. It is supervised by the identification label $t$ and the verification label $s$. The pre-trained CNN model can be CaffeNet \cite{krizhevsky2012imagenet}, VGG16 \cite{simonyan2014very} or ResNet-50 \cite{he2016deep}, from which we have removed the final fully-connected (FC) layer. The re-ID performance of the three models is comprehensively evaluated in Section \ref{exp}. Here, we do not provide detailed descriptions of the architecture of the CNN models and only take CaffeNet as an example in the following subsections. The three optimization objectives include two identification losses and one verification loss. We use the final convolutional activations $f$ as the discriminative descriptor for person re-ID, which is directly supervised by three objectives.

\subsection{Identification Loss}
There are two CaffeNets in our architecture. They share weights and predict the two identity labels of the input image pair simultaneously. In order to fine-tune the network on a new dataset, we replace the final fully-connected layer (1,000-dim) of the pre-trained CNN model with a  convolutional layer. The number of the training identities in Market-1501 is 751. So this convolutional layer has $751$ kernels of size $1 \times 1 \times 4096$ connected to the output $f$ of CaffeNet and then we add a softmax unit to normalize the output. The size of the result tensor is $1\times1\times751$. The Rectified Linear Unit (ReLU) is not added after this convolution. Similar to conventional multi-class recognition approaches, we use the cross-entropy loss for identity prediction, which is 
\begin{align}
\hat{p} &= softmax(\theta_{I}\circ f), \\
\mbox{Identif}(f,t,\theta_I) &= \sum^{K}_{i=1} -p_i\log(\hat{p_i}).
\end{align}
Here $\circ$ denotes the convolutional operation. $f$ is a $1\times1\times4,096$ tensor, $t$ is the target class and $\theta_I$ denotes the parameters of the added convolutional layer. $\hat{p}$ is the predicted probability, 
$p_i$ is the target probability. $p_i = 0$ for all $i$ except $p_t = 1$. 

\setlength{\tabcolsep}{15pt}
\begin{table}
\begin{center}
\begin{tabular}{l|cc}
\hline
Method & mAP & rank-1\\
\hline
CaffeNet (V) & 22.47 & 41.24 \\
CaffeNet (I) & 26.79 & 50.89 \\
CaffeNet (I+V) & \textbf{39.61} & \textbf{62.14} \\
\hline
VGG16 (V) & 24.29 & 42.99 \\
VGG16 (I) & 38.27 & 65.02 \\
VGG16 (I+V) & \textbf{47.45} & \textbf{70.16} \\
\hline
ResNet-50 (V) & 44.94 & 64.58 \\
ResNet-50 (I) & 51.48 & 73.69 \\
ResNet-50 (I+V) & \textbf{59.87} & \textbf{79.51} \\
\hline
\end{tabular}
\end{center}
\caption{Results on Market1501 \cite{zheng2015scalable} by identification loss and verification loss individually and jointly. ``I'' and ``V'' denote the identification loss and verification loss, respectively.}
\label{table:vs}
\end{table}

\subsection{Verification Loss}
While some previous works contain a matching function in the intermediate layers \cite{li2014deepreid,varior2016gated,ahmed2015improved}, our work directly compares the high-level features $f_1,f_2$ for similarity estimation. The high-level feature from the fine-tuned CNN has shown a discriminative ability \cite{zheng2016person,zheng2016mars} and it is more compact than the activations in the intermediate layers. So in our model, the pedestrian descriptor $f_1,f_2$ in the identification model are directly supervised by the verification loss. As shown in Fig. \ref{fig:2}, we introduce a non-parametric layer called Square Layer to compare the high-level features. It takes two tensors as inputs and outputs one tensor after subtracting and squaring element-wisely. The Square Layer is denoted as $f_s = (f_1 - f_2)^2$, where $f_1,f_2$ are the 4,096-dim embeddings and $f_s$ is the output tensor of the Square Layer. 

We then add a convolutional layer and the softmax output function to embed the resulting tensor $f_s$ to a 2-dim vector ($\hat{q_1}$, $\hat{q_2}$) which represents the predicted probability of the two input images belonging to the same identity. $\hat{q_1} + \hat{q_2} = 1.$ The convolutional layer takes $f_s$ as input and filters it with $2$ kernels of size $1 \times 1 \times 4096$. The ReLU is not added after this convolution. We treat pedestrian verification as a binary classification problem and use the cross-entropy loss that is similar to the one in the identification loss, which is
\begin{align}
\hat{q} &= softmax(\theta_{S}\circ f_s),\\
\mbox{Verif}(f_1,f_2,s,\theta_S) &= \sum_{i=1}^2 -q_i\log(\hat{q_i}).
\end{align}
Here $f_1,f_2$ are the two tensors of size $1\times1\times4096$. $s$ is the target class (same/different), $\theta_S$ denotes the parameters of the added convolutional layer and $\hat{q}$ is the predicted probability. If the image pair depicts the same person, $q_1 = 1, q_2 = 0$; otherwise, $q_1 = 0, q_2 = 1$. 

\begin{figure*}[t]
\begin{center}
%\fbox{\rule{0pt}{2in} \rule{0.9\linewidth}{0pt}}
   \includegraphics[width=0.8\linewidth]{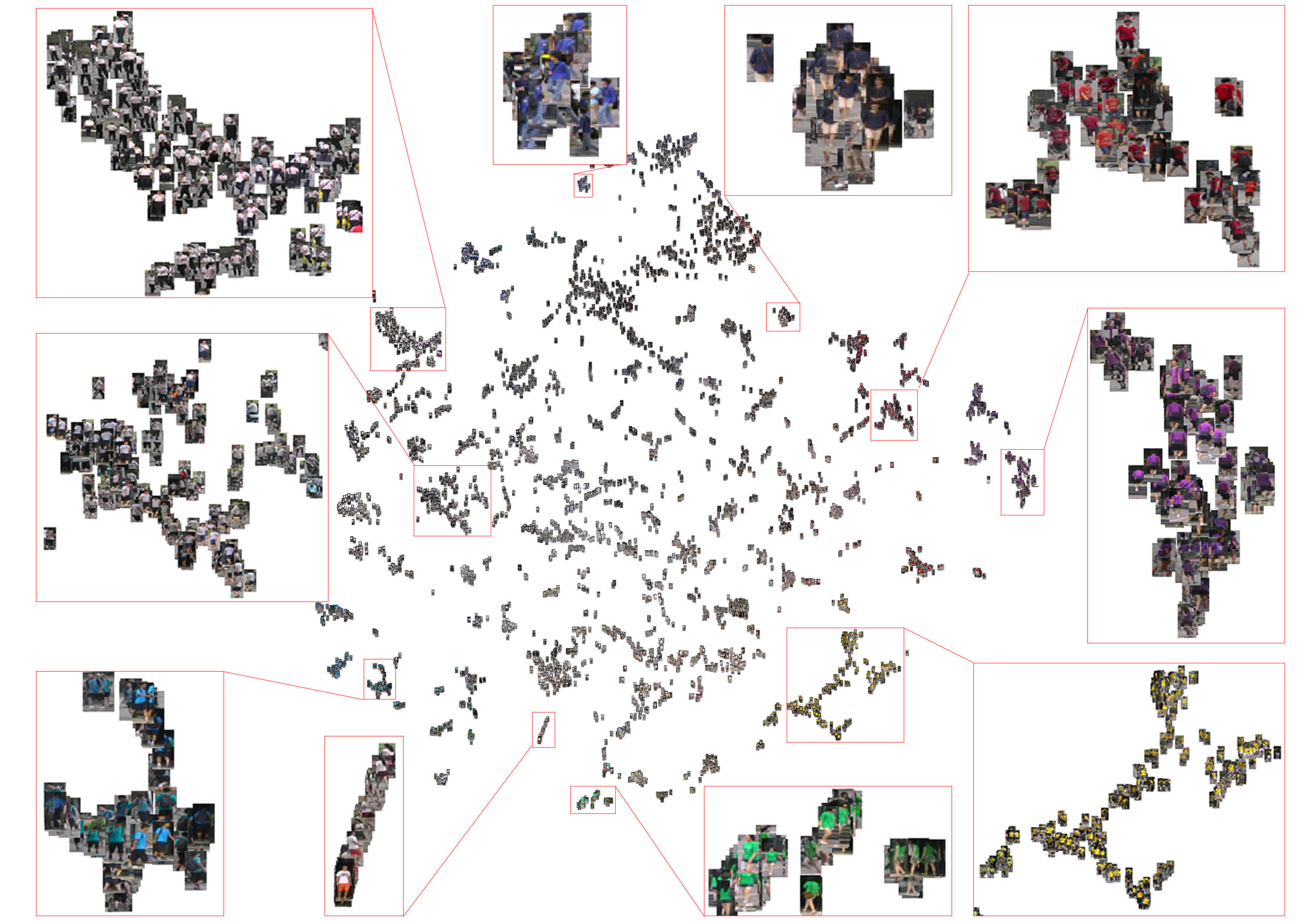}
\end{center}
   \caption{Barnes-Hut t-SNE visualization \cite{van2014accelerating} of our embedding on a test split (354 identity, 6868 images) of Market1501. Best viewed  when zoomed in. We find the color is the major clue for the person re-identification and our learned embedding is robust to some viewpoint variations.}
\label{fig:map}
\label{fig:onecol}
\end{figure*}

Departing from \cite{sun2014deep}, we do not use the contrastive loss \cite{hadsell2006dimensionality}. On the one hand, the contrastive loss, as a regression loss, forces the same-class embeddings to be as close as possible. It may make the model over-fitting because the number of training of each identity is limited in person re-ID. On the other hand, dropout \cite{srivastava2014dropout}, which introduces zero values at random locations, can not be applied on the embedding before the contrastive loss. But the cross-entropy loss in our model can work with dropout to regularize the model. In Section \ref{exp}, we show that the result using contrastive loss is 4.39\% and 6.55\% lower than the one using the cross-entropy loss on rank-1 accuracy and mAP respectively.

\subsection{Identification vs. Verification}
The proposed network is trained to minimize the three cross-entropy losses jointly. To figure out which objective contributes more, we train the identification model and verification model separately. Following the learning rate setting in Section \ref{3.6}, we train the models until convergence. We also train the network with the two losses jointly until two objectives both converge. As the quantitative results shown in Table \ref{table:vs}, the fine-tuned CNN model with two kinds of losses outperforms the one trained individually. This result has been confirmed on the three different network structures.

\begin{figure}[t]
\begin{center}
%\fbox{\rule{0pt}{2in} \rule{0.9\linewidth}{0pt}}
   \includegraphics[width=1\linewidth]{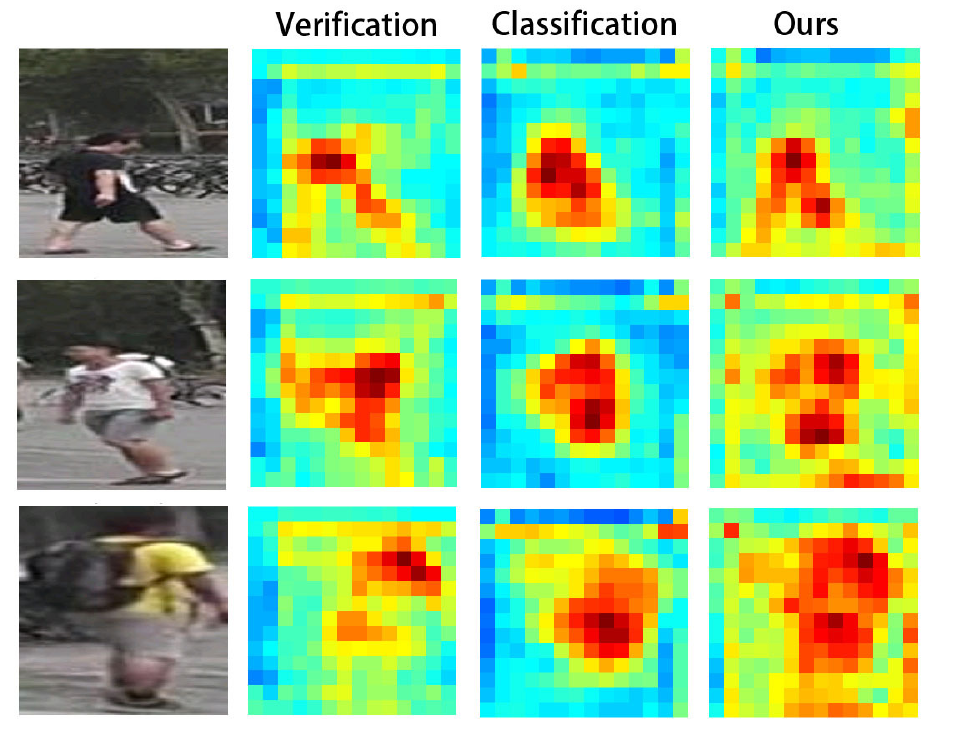}
\end{center}
   \caption{ Visualization of the activation maps in the ResNet-50 \cite{he2016deep} model trained by the two losses. The identification and the verification networks exhibit different activation patterns to the pedestrian. The proposed model takes advantages of both networks and the new activation map is almost a union of the two individual maps. Our model activates more neurons.}
\label{fig:4}
\label{fig:onecol}
\end{figure}

Further, we visualize the intermediate feature maps that are trained using ResNet-50 \cite{he2016deep} as the pretrained model and try to find the differences between identification loss and verification loss. We select three test images in the Market1501. One image is considered to be well detected and the other two images are not well aligned. Given one image as input, we get its activation in the intermediate layer ``res4fx'', the size of which is $14\times14$.  We visualize the sum of several activation maps. As shown in Fig. \ref{fig:4}, the identification and the verification networks exhibit different activation patterns to the pedestrian.  We find that if we use only one kind of loss, the network tends to find one discriminative part. The proposed model takes advantages of both networks, so the new activation map is mostly a union of the two individual maps. This also illustrates the complementary nature of the two baseline networks. The proposed model makes more neurons activated.  

Moreover, as shown in Fig. \ref{fig:map} we visualize  the  embedding by plot them to the 2-dimension map. In regard to Fig. \ref{fig:4}, we find the network usually has strong attention on the center part of the human (usually clothes) and it also illustrates the color of the clothes is the major clue for the person re-identification. 

\subsection{Training and Optimization} \label{3.6}

\textbf{Input preparation.} We resize all the training images to $256\times256$. The mean image computed from all the training images is subtracted from all the images. During training, all the images are randomly cropped to $227\times227$ for CaffeNet \cite{krizhevsky2012imagenet} and mirrored horizontally. For ResNet-50 \cite{he2016deep} and VGG16 \cite{simonyan2014very}, we randomly crop images to $224\times224$. We shuffle the dataset and use a random order of the images. Then we sample another image from the same/different class to compose a positive/negative pair. The initial ratio between negative pairs and positive pairs is $1:1$ to alleviate the prediction bias and we multiple it by a factor of $1.01$ every epoch until it reaches $1:4$, since the number of positive pairs is so limited that the network risks over-fitting. 

\textbf{Training.} We use the Matconvnet \cite{vedaldi15matconvnet} package for training and testing the embedding with CaffeNet \cite{krizhevsky2012imagenet}, VGG16 \cite{simonyan2014very} and ResNet-50 \cite{he2016deep}, respectively. The maximum number of training epochs is set to 75 for ResNet-50, 65 for VGG16net and 155 for CaffeNet. The batch size (in image pairs) is set to 128 for CaffeNet, 48 for VGG16 and ResNet-50. The learning rate is initialized as 0.001 and then set to 0.0001 for the final 5 epochs. We adopt the mini-batch stochastic gradient descent (SGD) to update the parameters of the network. There are three objectives in our network. Therefore, we first compute all the gradients produced by every objectives respectively and add the weighted gradients together to update the network. We assign a weight of 1 to the gradient produced by the verification loss and 0.5 for the two gradients produced by two identification losses. Moreover, we insert the dropout function \cite{srivastava2014dropout} before the final convolutional layer.

\textbf{Testing.} We adopt an efficient method to extract features as well as the activation in the intermediate layer. Because two CaffeNet share weights, our model has nearly the same memory consumption with the pretrained model. So we extract features by only activating one fine-tuned model. Given a $227 \times 227$ image, we feed forward the image to one CaffeNet in our network and obtain a 4,096-dim pedestrian descriptor $f$. Once the descriptors for the gallery sets are obtained, they are stored offline. Given a query image, its descriptor is extracted online. We sort the cosine distance between the query and all the gallery features to obtain the final ranking result. Note that the cosine distance is equivalent to Euclidean distance when the feature is L2-normalized.

\section{Experiments} \label{exp}
We mainly verify the proposed model on two large-scale datasets Market1501 \cite{zheng2015scalable} and CUHK03 \cite{li2014deepreid}. We report the results trained by three network structures. Besides, we also report the result on Market1501+500k dataset \cite{zheng2015scalable}. Meanwhile, the proposed architecture is also applied on the image retrieval task. We modify our model and test it on a popular image retrieval dataset, \ie  Oxford Buildings \cite{philbin2007object}. The performance is comparable to the state of the art.

\subsection{Dataset}
\textbf{Market1501} \cite{zheng2015scalable} contains 32,668 annotated bounding boxes of 1,501 identities. Images of each identity are captured by at most six cameras. According to the dataset setting, the training set contains 12,936 cropped images of 751 identities and testing set contains 19,732 cropped images of 750 identities and distractors. They are directly detected by the Deformable Part Model (DPM) instead of using hand-drawn bboxes, which is closer to the realistic setting. 
For each query, we aim to retrieve the ground truth images from the 19,732 candidate images.

The searching pool (gallery) is important to person re-identification. In the realistic setting, the scale of the gallery is usually large. The distractor dataset of Market1501 provides extra 500,000 bboxes, consisting of false alarms on the background as well as the persons not belonging to any of the original 1,501 identities \cite{zheng2015scalable}. When testing, we add the 500k images to the original gallery, which makes the retrieval more difficult.

\textbf{CUHK03} dataset \cite{li2014deepreid} contains 14,097 cropped images of 1,467 identities collected in the CUHK campus. Each identity is observed by two camera views and has 4.8 images in average for each view. The Author provides two kinds of bounding boxes. We evaluate our model on the bounding boxes detected by DPM, which is closer to the realistic setting. Following the setting of the dataset, the dataset is partitioned into a training set of 1,367 persons and a testing set of 100 persons. The experiment is repeated with 20 random splits. Both the single-shot and multiple-shot results will be reported.

\textbf{Oxford5k} buildings \cite{philbin2007object} consists of 5062 images collected from the internet and corresponding to particular Oxford landmarks. Some images have complex structures and may contain other buildings. The images corresponding to 11 Oxford landmarks are manually annotated and a set of 55 queries for 11 different landmarks are provided. This benchmark contains many high-resolution images and the mean image size of this dataset is $ 851 \times 921$.  

\setlength{\tabcolsep}{6pt}
\begin{table}
\begin{center}
\begin{tabular}{l|cc|cc}
%\small
\hline
\multirow{2}{*}{Method} & \multicolumn{2}{c|}{Single Query} & \multicolumn{2}{c}{Multi. Query}\\
& rank-1 & mAP & rank-1  & mAP \\
\hline
BoW + KISSME \cite{zheng2015scalable} & 44.42& 20.76  & - & -\\
SL \cite{chen2016similarity} & 51.90& 26.35  & - & -\\
Multiregion CNN \cite{ustinova2015multiregion} & 45.58& 26.11 & 56.59 & 32.26 \\
DADM \cite{su2016deep} & 39.4 & 19.6 & 49.0 & 25.8 \\ 
CAN \cite{liu2016end} & 48.24 & 24.43 & - & -\\
DNS \cite{zhang2016learning} & 55.43 & 29.87 & 71.56 & 46.03 \\
Fisher Network \cite{wu2016deep} & 48.15 & 29.94 & - & -\\
S-LSTM \cite{varior2016siamese} & - & - & 61.6 & 35.3\\
Gate Reid \cite{varior2016gated} & 65.88 & 39.55 & 76.04 & 48.45 \\
\hline\hline
CaffeNet-Basel. \cite{krizhevsky2012imagenet} & 50.89 & 26.79 & 59.80 & 36.50 \\
Ours(CaffeNet) & 62.14 & 39.61  & 72.21 & 49.62\\
\hline
VGG16-Basel. \cite{simonyan2014very} & 65.02 & 38.27 & 74.14 & 52.25 \\
Ours(VGG16) & 70.16 & 47.45 & 77.94 & 57.66 \\
\hline
ResNet-50-Basel. \cite{he2016deep} & 73.69 & 51.48 & 81.47 & 63.95 \\
Ours(ResNet-50) & \textbf{79.51} & \textbf{59.87} & \textbf{85.84} & \textbf{70.33} \\
\hline
\end{tabular}
\end{center}
\caption{Comparison with the state-of-the-art results on the Market1501 dataset. We also provide the results of the fine-tuned CNN baseline. The mAP and rank-1 precision are listed. SQ and MQ denote single query and multiply queries, respectively.}
\label{table:mr}
\end{table}
We use the rank-1 accuracy and mean average precision (mAP) for performance evaluation on Market1501 (+100k) and CUHK03, while on Oxford, we use mAP.

\subsection{Person Re-id Evaluation}

\textbf{Comparison with the CNN baseline.} We train the baseline networks according the conventional fine-tuning method \cite{zheng2016person,zheng2016survey}. The baseline networks are pretrained on ImageNet \cite{russakovsky2015imagenet} and fine-tuned to predict the person identities. As shown in Tab. \ref{table:mr}, we obtain 50.89\%, 65.02\% and 73.69\% rank-1 accuracy by CaffeNet \cite{krizhevsky2012imagenet}, VGG16 \cite{simonyan2014very} and ResNet-50 \cite{he2016deep}, respectively on Market1501. Note that using the baseline alone exceeds many previous works. Our model further improves these baselines on Market1501. The improvement can be observed on three network architectures. To be specific, we obtain 11.25\%, 5.14\% and 5.82\% improvement, respectively, using CaffeNet \cite{krizhevsky2012imagenet}, VGG16 \cite{simonyan2014very} and ResNet-50 \cite{he2016deep} on Market1501. Similarly, we observe 35.8\%, 49.1\% and 71.5\% baseline rank-1 accuracy on CUHK03 in single-shot setting. As show in Tab. \ref{table:single}, these baseline results exceed some previous works as well. We further get 14.0\%, 22.7\% and 11.9\% improvement on the baseline by our method.

These results show that our method can work with different networks and improve their results. It indicates that the proposed model helps the network to learn more discriminative features.  

\textbf{Cross-entropy vs. Contrastive loss.} We replace the cross-entropy loss with the contrastive loss as used in ``DeepID network''. However, we find a 4.39\% and 6.55\% drop in rank-1 and mAP. The ResNet-50 model using the contrastive loss has 75.12\% rank-1 accuracy and 53.32\% mAP. We speculate that the contrastive loss tends to over-fit on the re-ID dataset because no regularization is added to the verification. Cross-entropy loss designed in our model can work with the dropout function and avoid the over-fitting.

\begin{table}
\begin{center}
\begin{tabular}{l|c|c|c|c}
\hline
Method & rank-1 & rank-5 & rank-10 & mAP\\
\hline
KISSME \cite{kostinger2012large} & 11.7 & 33.3 & 48.0 & -\\
DeepReID \cite{li2014deepreid} & 19.9 & 49.3 & 64.7 & -\\ 
BoW+HS \cite{zheng2015scalable} & 24.3 & - & - & -\\
LOMO+XQDA \cite{liao2015person} & 46.3 & 78.9 & 88.6 &-\\
SI-CI \cite{wang2016joint} & 52.2 & 84.3 & 94.8 &-\\
DNS \cite{zhang2016learning} & 54.7 & 80.1 & 88.3 &-\\
\hline\hline
CaffeNet-Basel. & 35.8 & 65.3 & 77.96 & 42.6 \\
Ours (CaffeNet) & 59.8 & 88.3 & 94.2 & 65.8\\
\hline
VGG16-Basel. & 49.1 & 78.4 & 87.2 & 55.7 \\
Ours (VGG16) & 71.8 & 93.0 & 97.1 & 76.5 \\
\hline
ResNet-50-Basel. & 71.5 & 91.5 & 95.9 & 75.8 \\
Ours (ResNet-50) & \textbf{83.4} & \textbf{97.1} & \textbf{98.7} & \textbf{86.4}\\
\hline
\end{tabular}
\end{center}
\caption{Comparison with the state-of-the-art results reported on the CUHK03 dataset using the single-shot setting. The mAP and rank-1 accuracy are listed.}
\label{table:single}
\end{table}
\textbf{Comparison with the state of the art.} 
As shown in Table \ref{table:mr}, we compare our method with other state-of-the-art algorithms in terms of mean average precision (mAP) and rank-1 accuracy on Market1501. We report the single-query as well as multiple-query evaluation results. Our model (CaffeNet) achieves 62.14\% rank-1 accuracy and 39.61\% mAP, which is comparable to the state of the art 65.88\% rank-1 accuracy and 39.55\% mAP \cite{varior2016gated}. Our model using ResNet-50 produces the best performance 79.51\% in rank-1 accuracy and 59.87\% in mAP, which outperforms other state-of-the-art algorithms.

For CUHK03, we evaluate our method in the single-shot setting as shown in Tab. \ref{table:single}. There is only one right image in the searching pool. In the evaluation, we randomly select 100 images from 100 identities under the other camera as gallery. The proposed model yields 83.4\% rank-1 and 86.4\% mAP and outperforms the state-of-the-art performance.

\begin{figure}[t]
\begin{center}
%\fbox{\rule{0pt}{2in} \rule{0.9\linewidth}{0pt}}
   \includegraphics[width=1\linewidth]{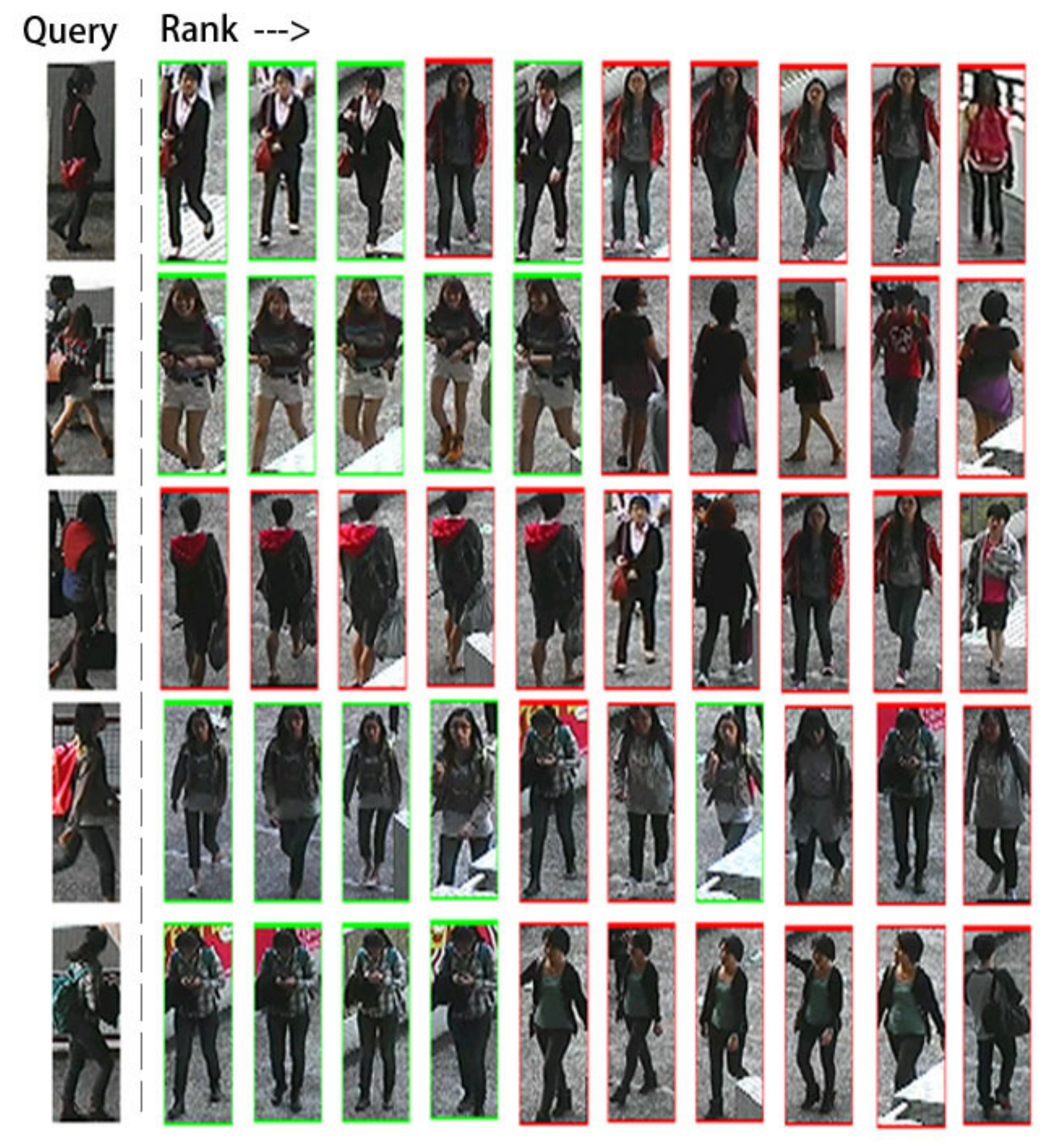}
\end{center}
   \caption{Pedestrian retrieval samples on CUHK03 dataset \cite{li2014deepreid} in the multi-shot setting. The images in the first column are the query images. The retrieval images are sorted according to the similarity scores from left to right.}
\label{fig:cuhk03}
\end{figure}

As shown in Tab. \ref{table:multi}, we also report the results in the multi-shot setting, which uses all the images from the other camera as gallery and the number of the gallery images is about 500. We think this setting is much closer to image retrieval and alleviate the unstable effect caused by the random searching pool under single-shot settings. Fig. \ref{fig:cuhk03} presents some re-ID samples on CUHK03 dataset. The images in the first column are the query images. The retrieval images are sorted according to the similarity scores from left to right. Most ground-truth candidate images are correctly retrieved. Although the model retrieves some incorrect candidates on the third row, we find it is a reasonable prediction since the man with red hat and blue coat is similar to the query. The proposed model yields 88.3\% rank-1 and 85.0\% mAP and also outperforms the state-of-the-art performance in the multi-shot setting.

\begin{table}
\begin{center}
\begin{tabular}{l|c|c|c|c}
\hline
Method & rank-1 & rank-5 & rank-10 & mAP\\
\hline
S-LSTM \cite{varior2016siamese} & 57.3 & 80.1 & 88.3 & 46.3 \\
Gate-SCNN \cite{varior2016gated} & 68.1 & 88.1 & 94.6 & 58.8\\
\hline\hline
CaffeNet-Basel. & 43.3 & 63.5 & 76.8 & 37.2 \\
Ours (CaffeNet) & 67.2 & 86.2 & 92.3 & 61.5 \\
\hline
VGG16-Basel. & 58.8 & 80.2 & 87.3 & 51.0 \\
Ours (VGG16) & 78.8 & 91.8 & 95.4 & 73.9 \\
\hline
ResNet-50-Basel. & 77.1 & 89.6 & 93.9 & 73.1 \\
Ours(ResNet-50) & \textbf{88.3} & \textbf{95.7} & \textbf{97.8} & \textbf{85.0}\\
\hline
\end{tabular}
\end{center}
\caption{Comparison with the state-of-the-art methods on the CUHK03 dataset under the multi-shot setting. The multi-shot setting uses the all images in the other camera as gallery. The mAP and rank-1 accuracy are listed.}
\label{table:multi}
\end{table}

\begin{figure}[t]
\begin{center}
%\fbox{\rule{0pt}{2in} \rule{0.9\linewidth}{0pt}}
   \includegraphics[width=1\linewidth]{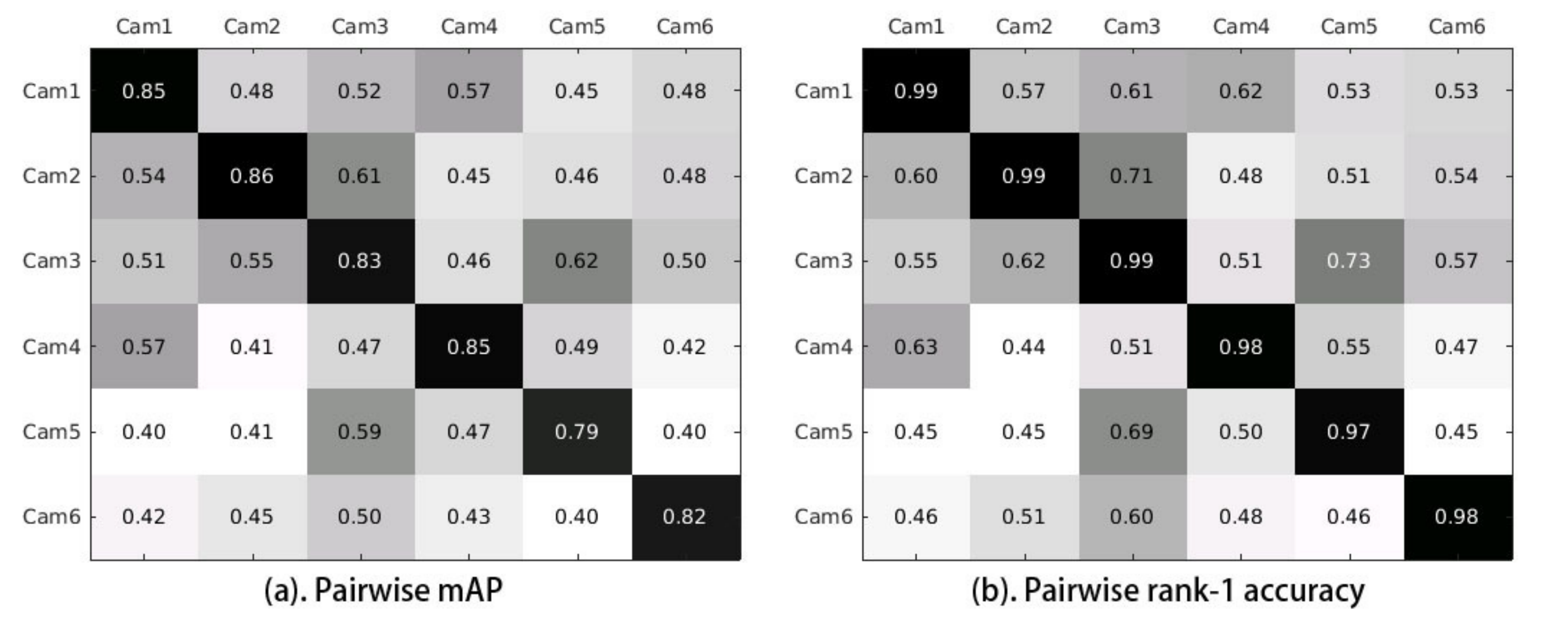}
\end{center}
   \caption{Re-identification performance between camera pairs on Market1501: (a) mAP and (b) rank-1 accuracy. Cameras on the vertical and horizontal axis correspond to the probe and gallery, respectively. The cross-camera average mAP and average rank-1 accuracy are 48.42\% and 54.42\%, respectively.}
\label{fig:camera}
\label{fig:onecol}
\end{figure}

\textbf{Results between camera pairs.} 
CUHK03 \cite{li2014deepreid} only contains two camera views. So this experiment is evaluated on Market1501 \cite{zheng2015scalable} since it contains six different cameras. We provide the re-identification results between all camera pairs in Fig. \ref{fig:camera}. Although camera-6 is a $720\times576$ low-resolution camera and captures distinct background with the other HD cameras, the re-ID accuracy between camera 6 and the others is relatively high. We also compute the cross-camera average mAP and average rank-1 accuracy: 48.42\% and 54.42\% respectively. Comparing to the previous reported results, \ie 10.51\% and 13.72\% in \cite{zheng2015scalable}, our method largely improves the performance and observes a smaller standard deviation between cameras. It suggests that the discriminatively learned embedding works under different viewpoints.

Further, Fig. \ref{fig:map} shows the Barnes-Hut t-SNE visualization \cite{van2014accelerating} on the learned embeddings of our model. By the clustering algorithm, the persons wearing the similar-color clothes are quit clustered together and are apart from other persons.  The learned pedestrian descriptor pay more attention to the color and it is  robust to some illusion and viewpoint variations. In realistic setting, we think color provides the most important information to figure out the person.

\begin{figure}[t]
\begin{center}
%\fbox{\rule{0pt}{2in} \rule{0.9\linewidth}{0pt}}
   \includegraphics[width=1\linewidth]{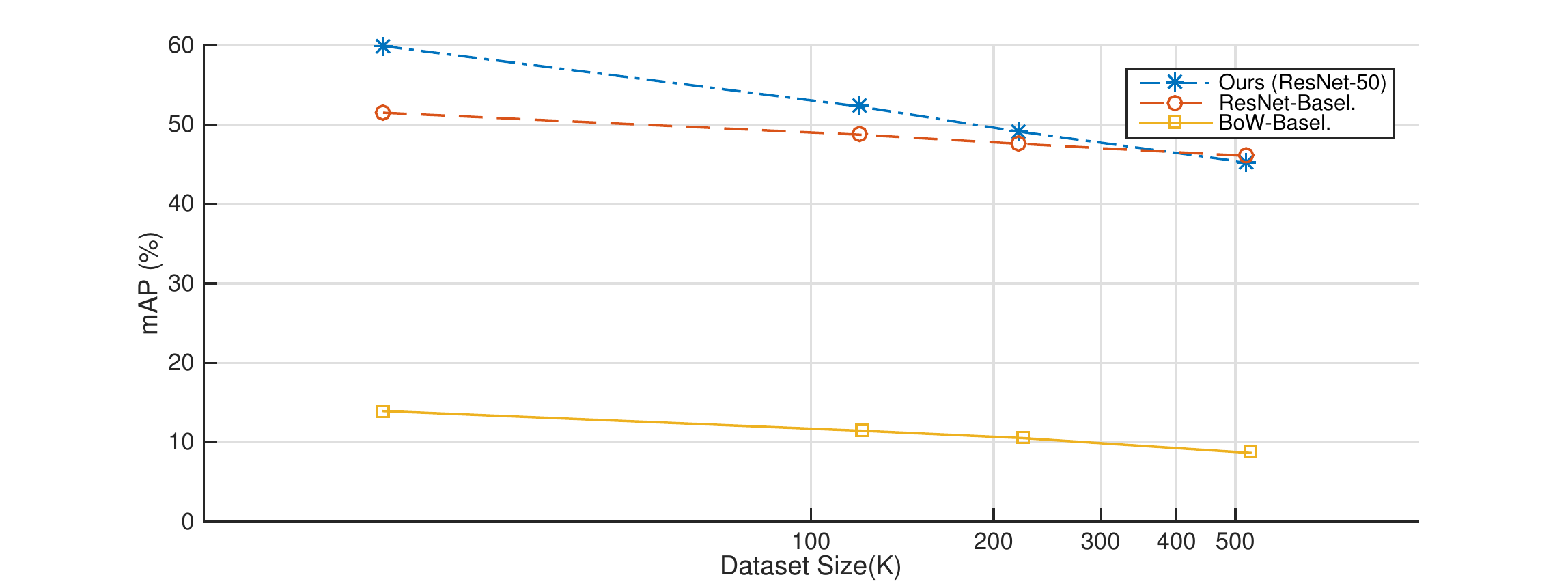}
\end{center}
   \caption{Impact of data size on Market1501+500K dataset. As the dataset gets larger, the accuracy drops.}
\label{fig:500k}
\end{figure}

\textbf{Large-scale experiments.}
The Market1501 dataset also provides an additional distractor set with 500k images  to enlarge the gallery. In general, more candidate images may confuse the image retrieval. The re-ID performance of our model (ResNet) on the large-scale dataset is presented in Tab. \ref{table:bigdata}. As the searching pool gets larger, the accuracy drops. With the gallery size of $500,000+19,732$, we still achieve 68.26\% rank1 accuracy and 45.24\% mAP. A relative drop 24.4\% from 59.87\% to 45.24\% on mAP is observed, compared to a relative drop 37.88\% from 13.94\% to 8.66\% in our previous work \cite{zheng2015scalable}. Besides, we also compare our result with the performance of the ResNet Baseline. As shown in Fig. \ref{fig:500k}, it is interesting that the re-ID precision of our model decreases more quickly comparing to the baseline model. We speculate that the Market1501 training set is relatively small in covering the pedestrian variations encountered in a much larger test set. In fact, the 500k dataset was collected in a different time (the same location) with the Market1501 dataset, so the transfer effect is large enough that the learned embedding is inferior to the baseline on the scale of 500 k images. In the future, we will look into this interesting problem and design more robust descriptors for the transfer dataset. 
\setlength{\tabcolsep}{5pt}
\begin{table}
\begin{center}
\footnotesize
\begin{tabular}{l|ccccc}
\hline
Method&Gallery size & 19,732 & 119,732 & 219,732 & 519,732\\
\hline
\multirow{2}{*}{ResNet Basel.}&rank-1 & 73.69 & 72.15 & 71.55 & 70.67\\ 
&mAP & 51.48 & 48.72 & 47.57 & 46.05\\
\hline
\multirow{2}{*}{Ours (ResNet)}&rank-1 & 79.51 & 73.78 & 71.50 & 68.26\\ 
&mAP & 59.87 & 52.28 & 49.11 & 45.24 \\
\hline
\end{tabular}
\end{center}
\caption{Impact of data size on Market1501+500K dataset. As the dataset gets larger, the accuracy drops.}
\label{table:bigdata}
\end{table}

\subsection{Instance Retrieval}
We apply the identification-verification model to the generic image retrieval task. Oxford5k \cite{philbin2007object} is a testing dataset containing buildings in the Oxford University. We train the network on another scene dataset proposed in \cite{radenovic2016cnn}, which comprises of a number of buildings without overlapping with the Oxford5k. Similarly, the model is trained to not only tell which building the image depicts but also determine whether the two input images are from the same architecture. The training data is high-resolution. In order to obtain more information from the high-resolution building images, we modify the final pooling layer of our model to a MAC layer \cite{tolias2015particular}, which outputs the maximum value over the whole activation map. This layer helps us to handle large images without resizing them to a fixed size and output a fixed-dimension feature to retrieve the images. During training, the input image is randomly cropped to $320 \times 320$ from $362 \times 362$ and mirrored horizontally. During testing, we keep the original size of the images that are not cropped or resized and extract the feature.

\begin{figure}[t]
\begin{center}
%\fbox{\rule{0pt}{2in} \rule{0.9\linewidth}{0pt}}
   \includegraphics[width=1\linewidth]{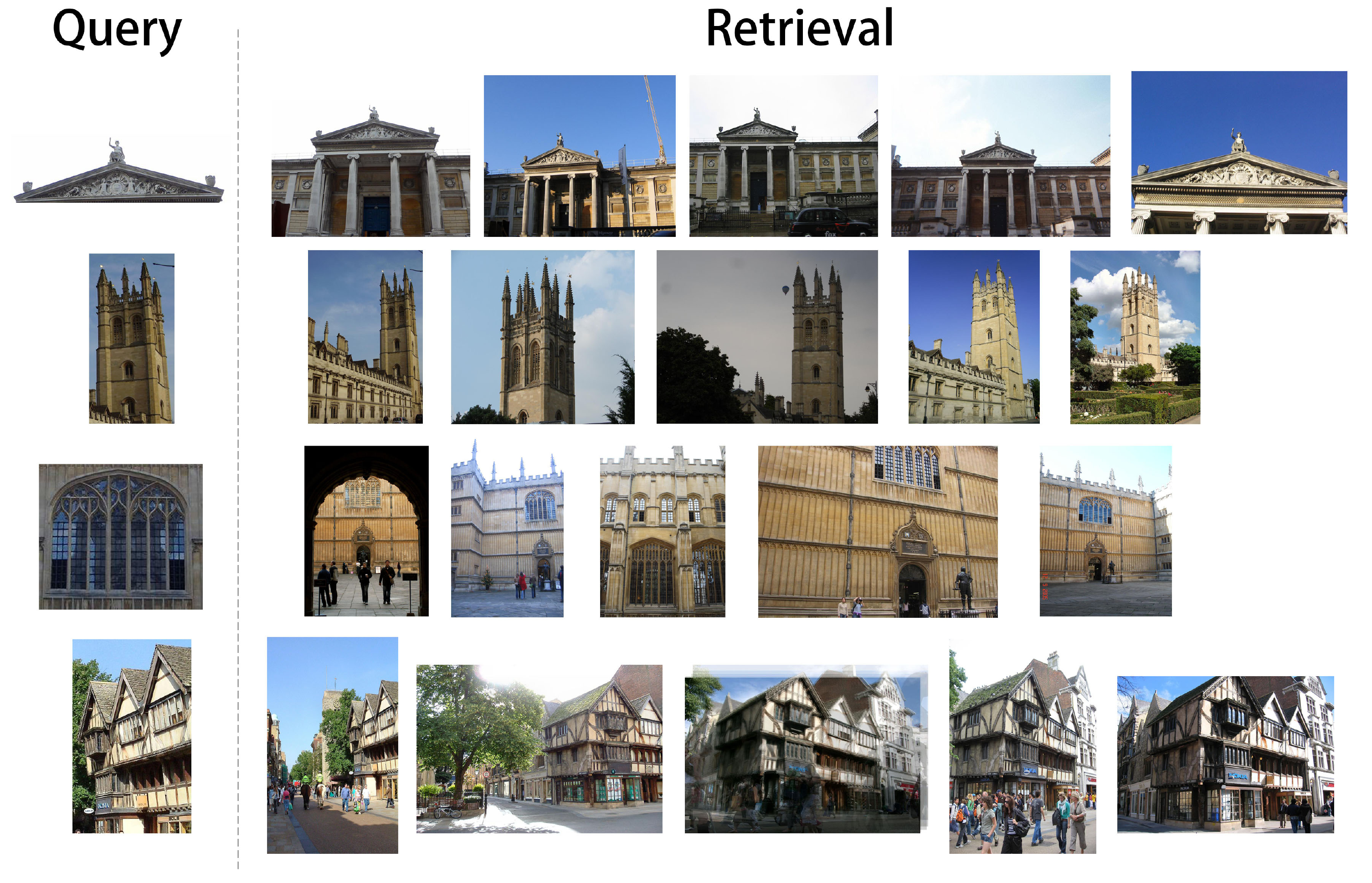}
\end{center}
   \caption{Example retrieval results on Oxford5k dataset \cite{philbin2007object} using the proposed embedding. The images in the first column are the query images. The retrieval images are sorted according to the similarity scores from left to right. The query images are usually from the part of the architectures.}
\label{fig:oxford5k}
\end{figure}

In Table \ref{table:oxford5k}, many previous works are based on CaffeNet or VGG16. For fair comparison, we report the baseline results and the results of our model based on these two network structures, respectively. Our model which uses CaffeNet as pretrained model outperforms the state of the art. Meanwhile, the model using VGG16 is comparable to the state-of-the-arts methods. The proposed method show a 6.0\% and 6.6\% improvement over the baseline networks CaffeNet and VGG16, respectively. We visualize some retrieval results in Fig. \ref{fig:oxford5k}. The images in the first column are the query images. The retrieval images are sorted according to the similarity scores from left to right. The main difficulty in the image retrieval is various object sizes in the image. In the first row, we use the roof (part of the building) to retrieve the images and the top five images are correct candidate images. The other retrieval samples also show our model is robust to the scale variations.

\begin{table}
\begin{center}
\begin{tabular}{l|c|c}
\hline
Method & CaffeNet mAP & VGG16 mAP\\
\hline
mVoc/BoW \cite{radenovic2015multiple} & 48.8 & -\\
CroW \cite{kalantidis2016cross} & - & 68.2\\
Neural codes \cite{babenko2014neural} & 55.7 & -\\
R-MAC \cite{tolias2015particular} & 56.1 & 66.9\\
R-MAC-Hard \cite{radenovic2016cnn} & 62.5 & 77.0\\
MAC-Hard(V) \cite{radenovic2016cnn} & 62.2 &\textbf{79.7} \\
\hline\hline
Finetuned-Baseline & 60.2 & 69.8\\
Ours & \textbf{66.2} & 76.4\\
\hline
\end{tabular}
\end{center}
\caption{Comparison of state-of-the-art results on the Oxford5k dataset. The mAP is listed. Results reported with the use of AlexNet\cite{krizhevsky2012imagenet} or VGGNet\cite{simonyan2014very} are marked by (A) or (V) respectively.}
\label{table:oxford5k}
\end{table}

\section{Conclusion}\label{sec:conclusion}
In this work, we propose a siamese network that simultaneously considers the identification loss and the verification loss. The proposed model learns a discriminative embedding and a similarity measurement at the same time. It outperforms the state of the art on two popular person re-ID benchmarks and shows potential ability to apply on the generic instance retrieval task. 

Future work includes exploring more novel applications of the proposed method, such as car recognition and fine-grained classification. Besides, we will investigate how to learn a robust descriptor to further improve the performance of the person re-identification on large-scale testing set.

% if have a single appendix:
%\appendix[Proof of the Zonklar Equations]
% or
%\appendix  % for no appendix heading
% do not use \section anymore after \appendix, only \section*
% is possibly needed

% use appendices with more than one appendix
% then use \section to start each appendix
% you must declare a \section before using any
% \subsection or using \label (\appendices by itself
% starts a section numbered zero.)
%

%\appendices
%\section{Proof of the First Zonklar Equation}
%Appendix one text goes here.

% you can choose not to have a title for an appendix
% if you want by leaving the argument blank
%\section{}
%Appendix two text goes here.
% use section* for acknowledgment
%\section*{Acknowledgment}
%The authors would like to thank...

% Can use something like this to put references on a page
% by themselves when using endfloat and the captionsoff option.
\ifCLASSOPTIONcaptionsoff
  \newpage
\fi

% trigger a \newpage just before the given reference
% number - used to balance the columns on the last page
% adjust value as needed - may need to be readjusted if
% the document is modified later
%\IEEEtriggeratref{8}
% The "triggered" command can be changed if desired:
%\IEEEtriggercmd{\enlargethispage{-5in}}

% references section
\bibliographystyle{IEEEtran}
\bibliography{IEEEabrv,mybib}

% biography section
% 
% If you have an EPS/PDF photo (graphicx package needed) extra braces are
% needed around the contents of the optional argument to biography to prevent
% the LaTeX parser from getting confused when it sees the complicated
% \includegraphics command within an optional argument. (You could create
% your own custom macro containing the \includegraphics command to make things
% simpler here.)
%\begin{IEEEbiography}[{\includegraphics[width=1in,height=1.25in,clip,keepaspectratio]{mshell}}]{Michael Shell}
% or if you just want to reserve a space for a photo:

% if you will not have a photo at all:

% You can push biographies down or up by placing
% a \vfill before or after them. The appropriate
% use of \vfill depends on what kind of text is
% on the last page and whether or not the columns
% are being equalized.

\vfill

% Can be used to pull up biographies so that the bottom of the last one
% is flush with the other column.
%\enlargethispage{-5in}

% that's all folks
\end{document}